# Keyphrase based Evaluation of Automatic Text Summarization

Fatma Elghannam
Electronics Research Institute
Cairo, Egypt

Tarek El-Shishtawy
Faculty of Computers and Information
Benha University, Benha, Egypt

## ABSTRACT
The development of methods to deal with the informative contents of the text units in the matching process is a major challenge in automatic summary evaluation systems that use fixed n-gram matching. The limitation causes inaccurate matching between units in a peer and reference summaries. The present study introduces a new Keyphrase based Summary Evaluator (KpEval) for evaluating automatic summaries. The KpEval relies on the keyphrases since they convey the most important concepts of a text. In the evaluation process, the keyphrases are used in their lemma form as the matching text unit. The system was applied to evaluate different summaries of Arabic multi-document data set presented at TAC2011. The results showed that the new evaluation technique correlates well with the known evaluation systems: Rouge-1, Rouge-2, Rouge-SU4, and AutoSummENG–MeMoG. KpEval has the strongest correlation with AutoSummENG–MeMoG, Pearson and spearman correlation coefficient measures are 0.8840, 0.9667 respectively.

## 1. INTRODUCTION
Evaluation of automatic text summarization systems using human evaluators requires expensive human efforts. This hard expensive effort has held up researchers looking for methods to evaluate summaries automatically. Current automated methods compare fragments of the summary to be assessed against one or more reference summaries (typically produced by humans), measuring how much fragments in the reference summary is present in the generated summary. One method is to consider the sentence as the fragment text unit in the evaluation process, but the problem is those sentences contain many individual pieces of information, which may not be used by humans for reference summaries. Choosing an appropriate fragment length and comparing it appropriately is a critical problem in the evaluation process. The problem is to extract the matching units that express the informative contents of a text. The misleading in choosing the informative content of a text, leads to unfortunate matching between two pieces of text in a peer and reference summaries.

Based on the intuition that the keyphrases represent the most important concepts of the text, we propose a Keyphrase based summary Evaluator (KpEval) for evaluating document summarization systems, considering the keyphrase as the matching text unit for the evaluation process.

KpEval idea is to count the matches between the peer summary and reference summaries for the essential parts of the summary text. KpEval have three main modules, i) lemma extractor module that breaks the text into words and extracts their lemma forms and the associated lexical and syntactic features, ii) keyphrase extractor that extracts important keyphrases in their lemma forms, and iii) the evaluator that scoring the summary based on counting the matched keyphrases occur between the peer summary and one or more reference summaries. The remaining of this paper is organized as follows: Section 2 reviews the previous works; Section 3 the proposed keyphrase based summary evaluator; Section 4 discusses the performance evaluation; and section 5 is the conclusion.

## 2. PREVIOUS WORKS
Evaluating summaries and automatic text summarization systems is not a straightforward process. Evaluation of automatic text summarization systems can be extrinsic or intrinsic evaluation methods [6]. In extrinsic evaluation, the summary quality is judged on the basis of how helpful summaries are for a given task. Intrinsic evaluation has mainly focused on the informativeness and coherence of summaries. This is often done by comparing the peer summary to reference/human summary. Many systems have been developed for automatic evaluation of the summary systems. Bleu (Bilingual Evaluation Understudy) [9] is an n-gram precision based evaluator metric initially designed for the evaluation of machine translation. The main idea of BLEU is to measure the translation closeness between a candidate translation and a set of reference translations with a numerical metric. They use a weighted average of variable length n-gram matches between system translations and a set of human reference translations. Lin et al. [8] have applied the same idea of Bleu to the evaluation of summaries. They used automatically computed accumulative n-gram matching scores between peer summaries and reference summaries as a performance indicator. ROUGE [9] stands for Recall-Oriented Understudy for Gisting Evaluation is a recall measure that counts the number of overlapping n-gram units between the peer summary generated by computer and several reference summaries. ROUGE has proved to be a successful algorithm. Several variants of the measure were introduced, such as ROUGE-N, ROUGE-S and ROUGE-SU. ROUGE-N is an n-gram recall between a candidate summary and a set of reference summaries. ROUGE-S is Skip-Bigram Co-Occurrence Statistics. Skip-bigram is any pair of words in their sentence order, allowing for arbitrary gaps. ROUGE-S, measures the overlap ratio of skip-bigrams between a candidate summary and a set of reference summaries. One potential problem for ROUGE-S is that it does not give any credit to a candidate sentence if the sentence does not have any word pair co-occurring with its references. The problem is solved by extending ROUGE-S with the addition of unigram as counting unit. The extended version is called ROUGE-SU. AutoSummENG–MeMoG (MeMoG) [3] is a summarization evaluation method that evaluates summaries by extracting and comparing graphs of character n-grams between the generated and model summaries. Hovy et al. [4] developed (BE) method, BE is a very small syntactic unit of content. They defined BE as: i) the head of a major syntactic constituent (noun, verb, adjective or adverbial phrases), expressed as a single item, or ii) a relation between a head-BE and a single dependent, expressed as a triple (head | modifier | relation). Their idea is to decompose reference and system summaries to lists of (BEs) units and then compare the two lists to obtain a similarity score. They include a syntactic parser to produce a parse tree and a set of 'cutting rules' to extract just the valid BEs from the tree. A modified version of BE is BEwT-E uses a set of transformations to match lexically different BEs that convey similar semantic content [5].

## 3. KEYPHRASE BASED SUMMARY EVALUATOR

A problem with methods using fixed n-gram matching is that they rely only on surface-level features, and the nonexistence of deep features that express the informative contents of the matching units [10]. Neglecting the linguistic features in the matching units misleads the matching process. We define two major types of errors that can occur between the matched units in a peer and reference summaries:

1) Under-matching, where non identical form units that cover the same concept are considered as unmatched units. This problem can occur at any of the NLP levels (lexical, syntactic, and semantic). For example the units (stages of education, education stages, education levels) convey the same concepts but with different syntactic structure or synonyms. Recognizing this problem needs different NLP analysis levels.

2) Over-matching, where the matched units does not reflect a real agreement in the concept between the peer summary and reference summary. For example matching the word "big" that exists in two different phrases in peer and reference summary like "big factories" and "big fish" is unfair as there is no real concept agreement.

Regarding this problem, attention must be paid to the informative content in the matching units. Based on the intuition that keyphrases represent the most important concepts of the text, in the proposed KpEval, keyphrases are considered as the matching units for the evaluation process. KpEval idea is to count the matches that occur between the peer summary and one or more reference summaries for the essential parts of the summary text (keyphrases). We adopted the existing lemma based keyphrase extractor LBAKE module [2] which is based on statistical techniques in addition to linguistic knowledge to extract the candidate keyphrases. KpEval technique starts by extracting the keyphrases for both the peer and reference summaries. For a peer summary, the number of matched keyphrases that occur with those existing in the reference summaries are counted. Precision, recall, and F-measure are calculated to measure the peer summary performance.

## 3.1 Features of the Keyphrase Based Summary Evaluator

- KpEval is based on counting the matched keyphrases that occur between the peer and reference summaries.

- The important keyphrases are extracted based on statistical and linguistic features. The existing LBAKE module is adopted for this purpose.

- Syntactic rules are used to identify the most informative phrases in a summary.

- The matching process is applied to keyphrases represented in their lemma form. So, different word forms that have the same meaning in a peer and reference summaries can be considered the same. This can be useful to overcome the lexical phase of the under_matching problem. For example, a word can have different plural forms in Arabic. So, the lemma forms of the two phrases "أعداد الطلاب", "عدد الطلبة" will be matched. To the best of our knowledge, none of the existing summary evaluation systems support an Arabic lemmatizer. Hovy et al. [4] extracts the basic elements (BE) – which are used in the matching process- based on the words syntactic feature.

- Precision, recall, and F-measure are used to evaluate the summary performance.

## 3.2 Steps of the Keyphrase Based Summary Evaluator

KpEval process has the following steps:

1. Extract the indicative keyphrases at lemma level using LBAKE module for both of peer and reference human summaries.

2. Count the matched keyphrases lemma forms that occur between the peer summary and each one of the reference summaries.

3. Calculate precision, recall, and F-measure to measure the peer summary performance.

### 3.2.1 Keyphrase Extraction

The first step is to pass the peer and reference summaries to the keyphrase extractor LBAKE module to extract the indicative keyphrases at lemma level. LBAKE is a supervised learning module for extracting keyphrases of single Arabic document. It is based on three main steps: Linguistic processing, candidate phrase extraction, and feature vector calculation. It starts by breaking the text into words and extracting their lemma forms and the associated lexical and syntactic features using the Arabic Lemmatizer [1]. And then the extractor extracts the keyphrases in their lemma form for both of the peer and reference summaries. The extractor is supplied with linguistic knowledge as well as statistical information. All possible phrases of one, two, or three consecutive words that appear in a given document are generated as n-gram terms. These n-gram words are accepted as a candidate keyphrases if they follow the following syntactic rules.

1- The candidate phrase can start only with some sort of nouns provided that not to be an adjective like general-noun, defined-noun, undefined noun, copulative noun and proper-noun.

2- The candidate phrase can end only with general-noun, place-noun, proper-noun, declined-noun, time-noun, augmented-noun, and adjective.

3- For three words phrase, the second word is allowed only to be a preposition, in addition to those cited in rule 2.

It is worthwhile to note that the rules applied are language-dependent, and the given rules are applicable only to Arabic language.

The importance of a keyphrase within a document is evaluated based on eight features:

- Number of words in each phrase.
- Frequency of the candidate phrase.
- Frequency of the most frequent single word in a candidate phrase.
- Location of the phrase sentence within the document.
- Location of the candidate phrase within its sentence.
- Relative phrase length to its containing sentence.
- Assessment of the phrase sentence verb content.
- Assessment as to whether the phrase sentence is in the form of a question.

Weights of these features were learned during building the classifier. The output of LBAKE is a set of keyphrases in their lemma form representing the input document.

The following example shows the generated keyphrases for two phrases a and b that exist in a peer and reference summary:

$a$: (" المعاهد العالية بالقرى " , " The high institutions in the villages")

$b$: (" الأبراج العالية بالقرى " ," The high towers in the villages")

After applying the filtering syntactic rules, the system will produce the n-gram keyphrases as illustrated in table 1.

**Table 1: Extracted keyphrases for the two phrases *a,b*.**

| KP (*a*) | KP (*b*) |
|---|---|
| المعاهد | الابراج |
| المعاهد العالية | الابراج العالية |
| المعاهد العالية بالقرى | الابراج العالية بالقرى |
| **القرى** | **القرى** |

Accordingly, only one matched keyphrase lemma form (قرية village) occurs between the two phrases. We mentioned here that according to the previous syntactic rules, units such as (العالية high), (العالية بالقرى high in the villages) that occur on both of the two phrases are not extracted as keyphrases, and consequently does not considered as an equivalent matched units. Note that the word sequence in the adjective phrase in Arabic is different from the English; the noun comes first then followed by the adjective. On the other hand, for an evaluation system that relies only on n-gram matching, the system would count (العالية بالقرى, العالية) as two extra matched units, regardless the different tenor speech in the two phrases. Using such syntactic rules in extracting keyphrases contributes well in assigning the most informative units, and at the same time reduces improper matching that can be occur in the evaluation process.

*3.2.2 Precision, Recall, and F measure calculation*

KpEval technique is based on evaluating precision, recall, and F-measure between the peer and reference summaries using the extracted keyphrases. For a peer summary, the number of overlapping keyphrases with each one of the reference summaries is calculated. Precision P, recall R, and F-measure are then evaluated using the following formulas [7]:

$$recall = \frac{\sum_{s \in (referencesummaries)} \sum_{kp \in s} count_{match}(KP)}{\sum_{s \in (referencesummaries)} \sum_{kp \in s} count\_ref(KP)}$$

$$precision = \frac{\sum_{s \in (referencesummaries)} \sum_{kp \in s} count_{match}(KP)}{\sum_{kp \in s} count\_sys(KP) \times no\_referencesummaries}$$

$$F = \frac{2 \times P \times R}{(P+R)}$$

Where $count\_ref(KP)$, $count\_sys(KP)$ is the number of keyphrases in their lemma form in the reference summaries and peer (system) summary respectively. $count_{match}(KP)$ is the maximum number of keyphrases co-occurring in a peer summary and in reference summaries. $no\_referencesummaries$ is the number of reference summaries.

## 4. PERFORMANCE EVALUATION

To evaluate the performance of the proposed system we compared it against other standard systems. We apply KpEval to evaluate a set of different participating systems at TAC 2011 MultiLing pilot, and compare the evaluation results against the existing results of Rouge-1, Rouge-2, Rouge-SU4, and AutoSummENG–MeMoG evaluation scores.

### 4.1 Data Set

The well-known TAC 2011 MultiLing Pilot 2011 Dataset[1] is used; the package contains all the dataset files related to the MultiLing 2011 Pilot. The data includes the peer summaries, human summaries, and results of Rouge-1, Rouge-2, Rouge-SU4, and AutoSummENG–MeMoG evaluation scores. The data set available in 7 languages including Arabic. We apply our test on the Arabic documents. For the Arabic language, there were seven participants (peers) in addition to the two baseline systems, for a total of nine results. The source texts contain ten collections of related newswire and newspaper articles. Each collection contains ten of related articles. The MultiLing task requires the participant to generate a single summary for each collection. The human summaries include three human (reference) summaries for each collection.

### 4.2 Evaluation Results

Table 2 illustrates KpEval average evaluation scores for the set of summarization systems participated at TAC 2011. We compared our summary performance results against four other systems: Rouge-1, Rouge-2, Rouge-SU4, and AutoSummENG–MeMoG. Pearson correlation coefficient is used to measure agreement with the scores, and the Spearman coefficient to measure correlation with the rankings. The results showed that KpEval correlates with the other four techniques; MeMoG has the strongest correlation with KpEval in both measures (0.8840 and 0.9667) as illustrated in Table 3.

**Table 2: Average scores by KpEval**

| SysID | KpEval |
|---|---|
| ID1 | 0.22761 |
| ID10 | 0.43185 |
| ID2 | 0.32429 |
| ID3 | 0.36593 |
| ID4 | 0.37414 |
| ID6 | 0.31322 |
| ID7 | 0.18409 |

---

[1] http://www.nist.gov/tac/2011/Summarization/

| | |
|---|---|
| ID8 | 0.26476 |
| ID9 | 0.23494 |

Figure1 shows the participating systems superiority ranking assessed by the different four evaluation techniques. The experiment shows that KpEval has almost agreement with different evaluation techniques for assessing the participating systems superiority ranking.

**Table 3: Pearson and spearman correlation coefficient between KpEval and other systems**

| | R1 | R2 | R-SU4 | MEMOG |
|---|---|---|---|---|
| **Pearson** | 0.8824 | 0.7487 | 0.8133 | 0.8840 |
| **Spearman** | 0.8167 | 0.6333 | 0.7667 | 0.9667 |

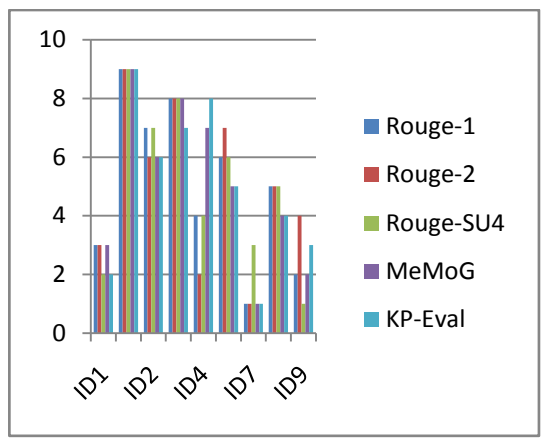

**Figure1 : Participating summarization systems superiority ranking by different evaluation techniques**

## 5. CONCLUSION AND FUTURE WORK

In this paper we introduced a keyphrase based evaluation system KpEval for assessing automatic summaries. KpEval is based on counting the matched keyphrases lemma form of the summary to be assessed against reference summaries. KpEval has three main steps i) extract the keyphrases for both peer and reference summaries ii) count the matched keyphrases occurring between the peer and each one of the reference summaries, and iii) calculate precision, recall, and F-measure. To measure the validity of the new system, Pearson and spearman correlation coefficient measures were tested between the results of KpEval against other evaluation systems: Rouge-1, Rouge-2, Rouge-SU4, and MeMoG using TAC 2011 dataset. The results showed that KpEval correlates with the four techniques. MeMoG has the strongest correlation with KpEval , Pearson and spearman measures are 0.8840, and 0.9667 respectively. Feature work includes testing the proposed technique for documents in other languages, especially Semitic languages.

## REFERENCES

[1] El-Shishtawy, T., & El-Ghannam, F. 2012. An Accurate Arabic Root-Based Lemmatizer for Information Retrieval Purposes. International Journal of Computer Science Issues (IJCSI), 9(1).

[2] El-Shishtawy, T., & El-Ghannam, F. 2012, May. Keyphrase based Arabic summarizer (KPAS). Informatics and Systems (INFOS), 2012 8th International Conference. (pp. NLP-7. IEEE).

[3] Giannakopoulos, G., Karkaletsis, V., Vouros, G., & Stamatopoulos, P. 2008. Summarization system evaluation revisited: N-gram graphs. ACM Transactions on Speech and Language Processing (TSLP), 5(3), 5.

[4] Hovy, E., Lin, C. Y., & Zhou, L. 2005. Evaluating duc 2005 using basic elements. In Proceedings of DUC (Vol. 2005).

[5] Hovy, E., Lin, C. Y., Zhou, L., & Fukumoto, J. 2006, May. Automated summarization evaluation with basic elements. In Proceedings of the Fifth Conference on Language Resources and Evaluation (LREC 2006) (pp. 604-611).

[6] Jones, K. S., & Galliers, J. R. (Eds.). 1996. Evaluating natural language processing systems: An analysis and review (Vol. 1083). Springer.

[7] Lin, C. Y. 2004, July. Rouge: A package for automatic evaluation of summaries. In Text Summarization Branches Out: Proceedings of the ACL-04 Workshop (pp. 74-81).

[8] Lin, C. Y., & Hovy, E. 2003, May. Automatic evaluation of summaries using n-gram co-occurrence statistics. In Proceedings of the 2003 Conference of the North American Chapter of the Association for Computational Linguistics on Human Language Technology-Volume 1 (pp. 71-78). Association for Computational Linguistics.

[9] Papineni, K., Roukos, S., Ward, T., & Zhu, W. J. 2002, July. BLEU: a method for automatic evaluation of machine translation. In Proceedings of the 40th annual meeting on association for computational linguistics (pp. 311-318). Association for Computational Linguistics.

[10] Tratz, S., & Hovy, E. 2008. Summarization evaluation using transformed basic elements. In Proceedings of the 1st Text Analysis Conference.